\def\year{2020}\relax
\documentclass[letterpaper]{article} 
\usepackage{aaai20}  
\usepackage{times}  
\usepackage{helvet} 
\usepackage{courier}  
\usepackage[hyphens]{url}  
\usepackage{graphicx} 
\usepackage{graphicx}  
\usepackage{url}            
\usepackage{booktabs}       
\usepackage{amsfonts}       

\usepackage{amsmath}
\usepackage{makecell}
\usepackage{algorithm}
\usepackage{algorithmic}
\usepackage{graphicx}
\usepackage{caption,subcaption}
\renewcommand{\algorithmicrequire}{\textbf{Input:}}
\renewcommand{\algorithmicensure}{\textbf{Output:}}

\newcommand{\task}{\mathcal{T}}

\newcommand{\D}{\mathcal{D}}
\newcommand{\DTrain}{\mathcal{D}^{Train}}

\newcommand{\states}{\mathcal{S}}
\newcommand{\actions}{\mathcal{A}}
\newcommand{\argmax}{\mathop{{\arg\!\max}}}

\newcommand{\citet}[1]{\citeauthor{#1} \shortcite{#1}}
\newcommand{\citep}{\cite}

\usepackage{soul}
\usepackage{xcolor}

\title{Learning to Interactively Learn and Assist}
\author{Mark Woodward\thanks{Work done as a part of the Goolge AI Residency}, Chelsea Finn, Karol Hausman\\Google Brain, Mountain View\\\{markwoodward,chelseaf,karolhausman\}@google.com}
\begin{document}

\maketitle

\begin{abstract}
  When deploying autonomous agents in the real world, we need effective ways of communicating objectives to them. Traditional skill learning has revolved around reinforcement and imitation learning, each with rigid constraints on the format of information exchanged between the human and the agent. While scalar rewards carry little information, demonstrations require significant effort to provide and may carry more information than is necessary. Furthermore, rewards and demonstrations are often defined and collected before training begins, when the human is most uncertain about what information would help the agent. In contrast, when humans communicate objectives with each other, they make use of a large vocabulary of informative behaviors, including non-verbal communication, and often communicate throughout learning, responding to observed behavior. In this way, humans communicate intent with minimal effort. In this paper, we propose such \emph{interactive learning} as an alternative to reward or demonstration-driven learning. To accomplish this, we introduce a multi-agent training framework that enables an agent to learn from another agent who knows the current task. Through a series of experiments, we demonstrate the emergence of a variety of interactive learning behaviors, including information-sharing, information-seeking, and question-answering. Most importantly, we find that our approach produces an agent that is capable of learning interactively from a human user, without a set of explicit demonstrations or a reward function, and achieving significantly better performance cooperatively with a human than a human performing the task alone.
\end{abstract}

\section{Introduction}
\label{sec:introduction}

\begin{figure*}
    \centering
    \begin{subfigure}[t]{0.245\textwidth}
        \centering
        \captionsetup{width=0.99\linewidth}
	    \includegraphics[width=1.25in]{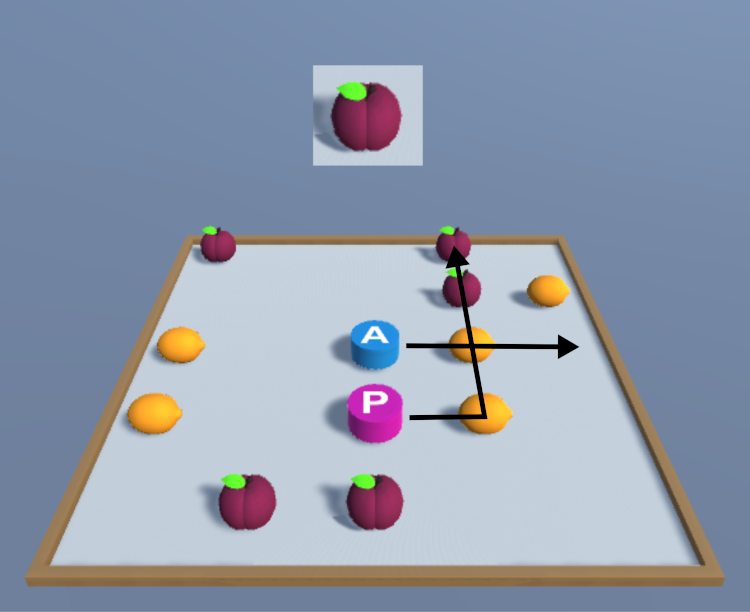}
	    \caption{\footnotesize After 100 gradient steps}
	    \label{fig:exp4_traces_100}
	\end{subfigure}
    \begin{subfigure}[t]{0.245\textwidth}
        \centering
        \captionsetup{width=0.99\linewidth}
	    \includegraphics[width=1.25in]{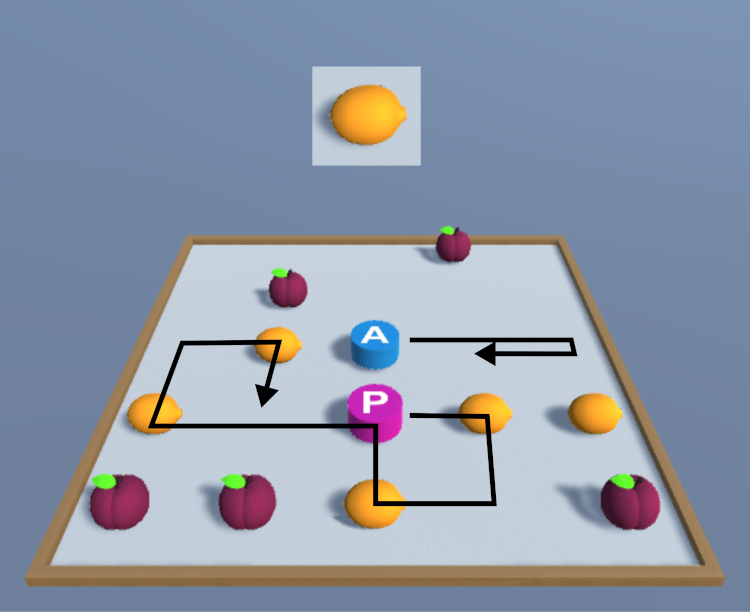}
	    \caption{\footnotesize After 1k gradient steps}
        \label{fig:exp4_traces_1k}
    \end{subfigure}
    \begin{subfigure}[t]{0.245\textwidth}
        \centering
        \captionsetup{width=0.99\linewidth}
	    \includegraphics[width=1.25in]{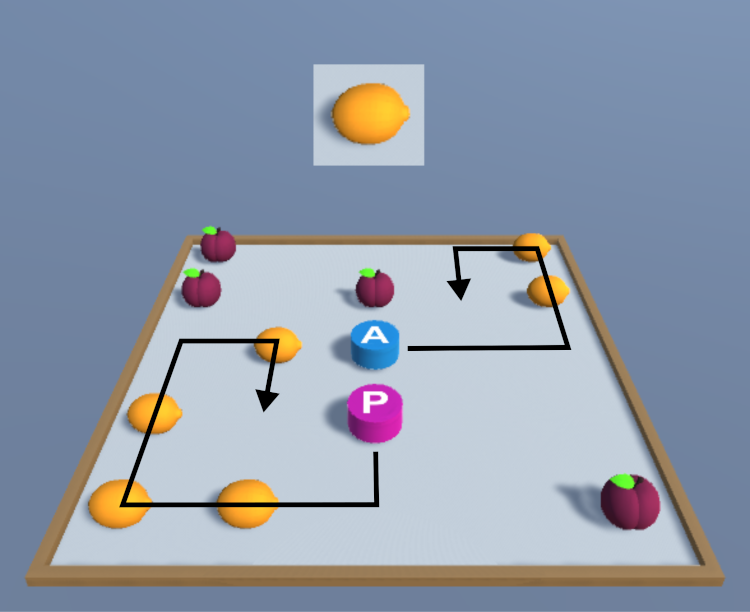}
        \caption{\footnotesize After 40k gradient steps}
	    \label{fig:exp4_traces_40k}
    \end{subfigure}
    \begin{subfigure}[t]{0.245\textwidth}
        \centering
        \captionsetup{width=0.99\linewidth}
	    \includegraphics[width=1.25in]{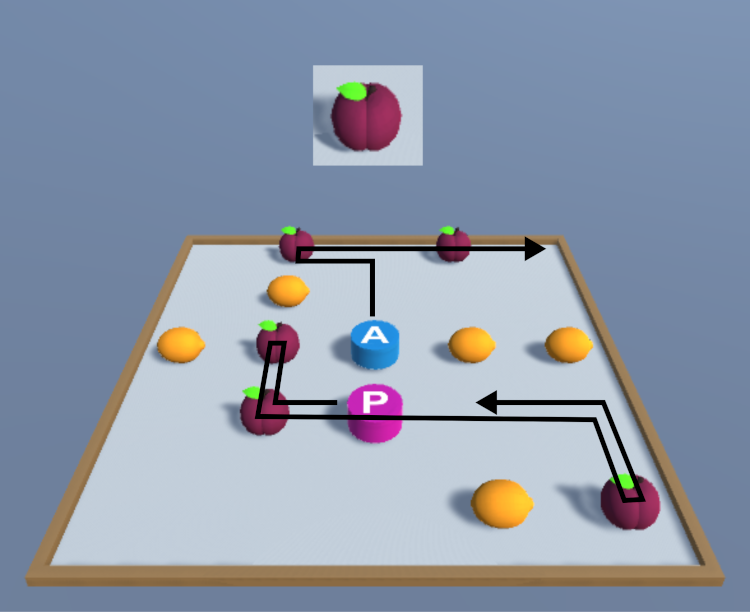}
        \caption{\footnotesize With human principal}
	    \label{fig:exp4_traces_human}
    \end{subfigure}
	\caption{\footnotesize Episode traces after 100, 1k, and 40k pre-training steps for the cooperative fruit collection domain of Experiment 4. The principal agent ``P'' (pink) is told the fruit to be collected, lemons or plums, in its observations. Within an episode, the assistant agent ``A'' (blue) must infer the fruit to be collected from observations of the principal. Each agent observes an overhead image of itself and its nearby surroundings.
	By the end of training (\subref{fig:exp4_traces_40k}) the assistant is inferring the correct fruit and the agents are coordinating. This inference and coordination transfers to human principals~(\subref{fig:exp4_traces_human}). An interactive game and videos for all experiments are available at: \url{https://interactive-learning.github.io}}
	\label{fig:exp4_traces}
\end{figure*}

Many tasks that we would like our agents to perform, such as unloading a dishwasher, straightening a room, or restocking shelves are inherently user-specific, requiring information from the user in order to fully learn all the intricacies of the task.
The traditional paradigm for agents to learn such tasks is through rewards and demonstrations.
However, iterative reward engineering with untrained human users is impractical in real-world settings, while demonstrations are often burdensome to provide.
In contrast, humans learn from a variety of interactive communicative behaviors, including nonverbal gestures and partial demonstrations, each with their own information capacity and effort.
Can we enable agents to learn tasks from humans through such unstructured interaction, requiring minimal effort from the human user?

The effort required by the human user is affected by many aspects of the learning problem, including restrictions on when the agent is allowed to act and restrictions on the behavior space of either human or agent, such as limiting the user feedback to rewards or demonstrations.
We consider a setting where both the user and the agent are allowed to act throughout learning, which we refer to as \emph{interactive learning}.
Unlike collecting a set of demonstrations before training, interactive learning allows the user to selectively act only when it deems the information is necessary and useful, reducing the user's effort.
Examples of such interactions include allowing user interventions or agent requests, for demonstrations~\cite{kelly2018hgd}, rewards~\cite{warnell2018dti,arumugam2019drl}, or preferences~\cite{christiano2017drl}.
While these methods allow the user to provide feedback throughout learning, the communication interface is restricted to structured forms of supervision, which may be inefficient for a given situation.
For example, in a dishwasher unloading task, given the history of learning, it may be sufficient to point at the correct drawer rather than provide a full demonstration. 

To this end, we propose to allow the agent and the user to exchange information through an unstructured interface. 
To do so, the agent and the user need a common prior understanding of the meaning of different unstructured interactions, along with the context of the space of tasks that the user cares about. 
Indeed, when humans communicate tasks to each other, they come in with rich prior knowledge and common sense about what the other person may want and how they may communicate that, enabling them to communicate concepts effectively and efficiently~\cite{peloquin2019irp}.

In this paper, we propose to allow the agent to acquire this prior knowledge through joint pre-training with another agent who knows the task and serves as a human surrogate. 
The agents are jointly trained on a variety of tasks, where actions and observations are restricted to the physical environment.
Since the first agent is available to assist, but only the second agent is aware of the task, interactive learning behaviors should emerge to accomplish the task efficiently.
We hypothesize that, by restricting the action and observation spaces to the physical environment, the emerged behaviors can transfer to learning from a human user. 
An added benefit of our framework is that, by training on a variety of tasks from the target task domain, much of the non-user specific task prior knowledge is pre-trained into the agent, further reducing the effort required by the user.

We evaluate various aspects of agents trained with our framework on several simulated object gathering task domains, including a domain with pixel observations, shown in Figure~\ref{fig:exp4_traces}.
We show that our trained agents exhibit emergent information-gathering behaviors in general and explicit question-asking behavior where appropriate.
Further, we conduct a user study with trained agents, where the users score significantly higher with the agent than without the agent, which demonstrates that our approach can produce agents that can learn from and assist human users.

The key contribution of our work is a training framework that allows agents to quickly learn new tasks from humans through unstructured interactions, without an explicitly-provided reward function or demonstrations. 
Critically, our experiments demonstrate that agents trained with our framework generalize to learning test tasks from human users, demonstrating interactive learning with a human in the loop.
In addition, we introduce a novel multi-agent model architecture for cooperative multi-agent training that exhibits improved training characteristics. 
Finally, our experiments on a series of object-gathering task domains illustrate a variety of emergent interactive learning behaviors and demonstrate that our method can scale to raw pixel observations.

\section{Related Work}
\label{sec:related-work}

The traditional means of passing task information to an agent include specifying a reward function~\citep{barto1998rli} 
that can be hand-crafted for the task~\citep{singh2009rewards,levine2016end,chebotar-hausman-zhang17icml} and providing demonstrations~\citep{schaal1999ilr,abbeel2004apprenticeship} before the agent starts training.
More recent works explore the concept of the human supervision being provided throughout training by either providing rewards during training~\citep{isbell2001cas,thomaz2005rti,warnell2018dti,dattari2018ilc} or demonstrations during training; either continuously~\citep{ross2011ari,kelly2018hgd} or at the agent's discretion~\citep{ross2011reduction,borsa2017olr,xu2018lpi,hester2018dqd,james2018tec,yu2018osi,krening2018naa,brown2018raa}.
In all of these cases, however, the reward and demonstrations are the sole means of interaction.

Another recent line of research involves the human expressing their preference between agent generated trajectories~\citep{christiano2017drl,mindermann2018air,ibarz2018rlh}.
Here again, the interaction is restricted to a single modality.

Our work builds upon the idea of meta-learning, or learning-to-learn~\citep{schmidhuber1987eps,bengio1991lsl,thrun2012learningtolearn}. Meta-learning for control has been considered in the context of reinforcement learning~\citep{duan2016rl2,wang2016learning,finn2017mam} and imitation learning~\citep{duan2017one,yu2018one}. 
Our problem setting differs from these, as the agent is learning by observing and interacting with another agent, as opposed to using reinforcement or imitation learning. 
In particular, our method builds upon recurrence-based meta-learning approaches~\citep{santoro2016osl,duan2016rl2,wang2016learning} in the context of the multi-agent task setting.

When a broader range of interactive behaviors is desired, prior works have introduced a multi-agent learning component~\citep{potter1994cca,palmer2018lma}. 
The following methods are closely related to ours in that, during training, they also maximize a joint reward function between the agents and emerge cooperative behavior~\citep{gupta2017cmc,foerster2018cma,foerster2016lcd,lazaridou2016mce,andreas2017tn}.
Multiple works~\cite{gupta2017cmc,foerster2018cma} emerge cooperative behavior but in task domains that do not require knowledge transfer between the agents, while others~\cite{foerster2016lcd,lazaridou2016mce,lowe2017maa,andreas2017tn,mordatch2018egc} all emerge communication over a communication channel. 
Such communication is known to be difficult to interpret~\citep{lazaridou2016mce}, without post-inspection~\citep{mordatch2018egc} or a method for translation~\citep{andreas2017tn}. Critically, none of these prior works conduct user experiments to evaluate transfer to humans.

\citet{mordatch2018egc} experiment with tasks similar to ours, in which information must be communicated between the agents, and communication is restricted to the physical environment. This work demonstrates the emergence of pointing, demonstrations, and pushing behavior. Unlike this prior approach, however, our algorithm does not require a differentiable environment. We also demonstrate our method with pixel observations and conduct a user experiment to evaluate transfer to humans.

~\citet{Laird2017itl} describe desiderata for interactive learning systems. Our method primarily addresses the desiderata of efficient interaction and accessible interaction.

\section{Preliminaries}
\label{sec:preliminaries}

In this section, we review the cooperative partially observable Markov game~\citep{littman1994mgf}, which serves as the foundation for tasks in Section~\ref{sec:methods}.
A cooperative partially observable Markov game is defined by the tuple $\langle$ $\states$, \{$\actions^i$\}, $T$, $\mathcal{R}$, \{$\Omega^i$\}, \{$O^i$\}, $\gamma$, $H$ $\rangle$, where $i\in\{1..N\}$ indexes the agent among $N$ agents,
$\states$, $\actions^i$, and $\Omega^i$ are state, action, and observation spaces, 
$T:\states \times \{\actions^i\} \times \states' \rightarrow \mathbb{R}$ is the transition function,
$\mathcal{R}:\states \times \{\actions^i\} \rightarrow \mathbb{R}$ is the reward function,
$O^i: \states \times \Omega^i \rightarrow \mathbb{R}$ are the observation functions,
$\gamma$ is the discount factor,
and $H$ is the horizon.

The functions $T$, $\mathcal{R}$, and $O^i$ are not accessible to the agents. At time $t$, the environment accepts actions $\{a_t^i\} \in \{\actions^i\}$, samples $s_{t+1} \sim T(s_t, \{a^i_t\})$, and returns reward $r_t \sim \mathcal{R}(s_t, \{a^i_t\})$ and observations $\{o^i_{t+1}\} \sim \{O^i(s_{t+1})\}$.
The objective of the game is to choose actions to maximize the expected discounted sum of future rewards:
\begin{equation} \label{eqn:pomdp_objective}
\argmax_{\{a^i_{t_0}|o^i_{t_0}\}} \mathop{{}\mathbb{E}}_{s,o^i,r} \big[ \sum_{t=t_0}^H \gamma^{t-t_0} r_t \big].
\end{equation}
Note that, while the action and observation spaces vary for the agents, they share a common reward which leads to a cooperative task.

\begin{figure}
    \centering
	\begin{subfigure}[b]{0.23\textwidth}
	    \centering
	    \includegraphics[width=1.6in]{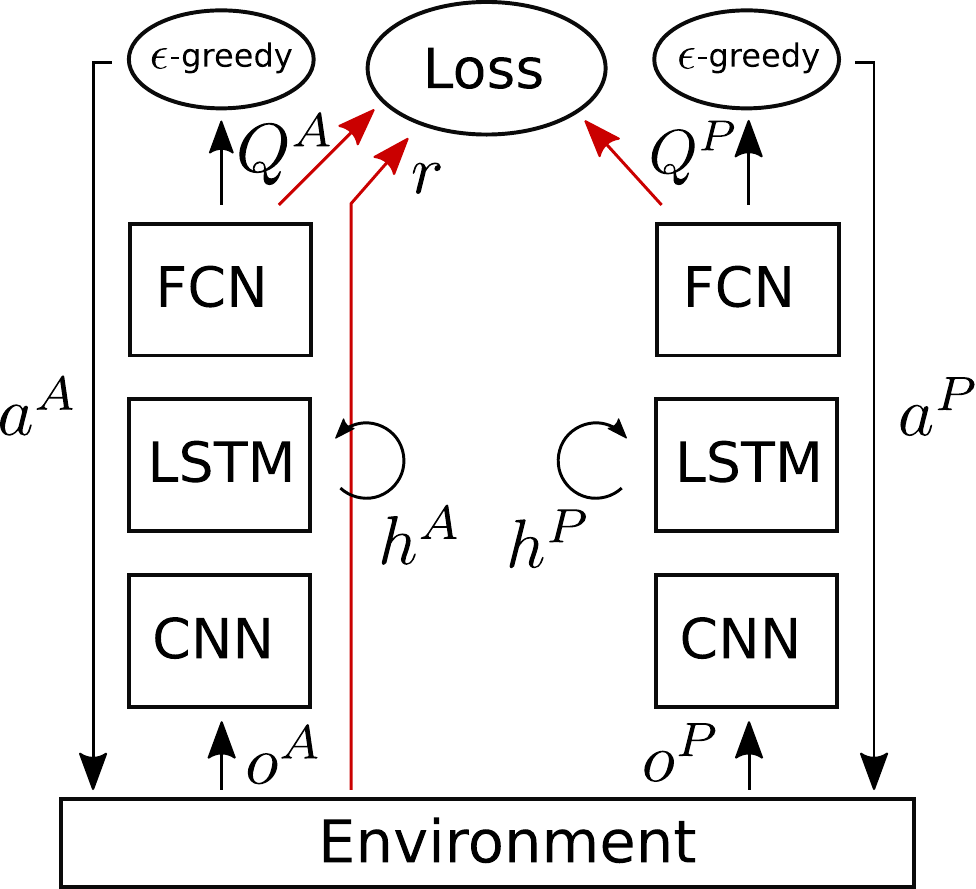}
	    \caption{\footnotesize MAIDRQN}
		\label{fig:maidrqn}
    \end{subfigure}
    \begin{subfigure}[b]{0.23\textwidth}
	    \centering
	    \includegraphics[width=1.6in]{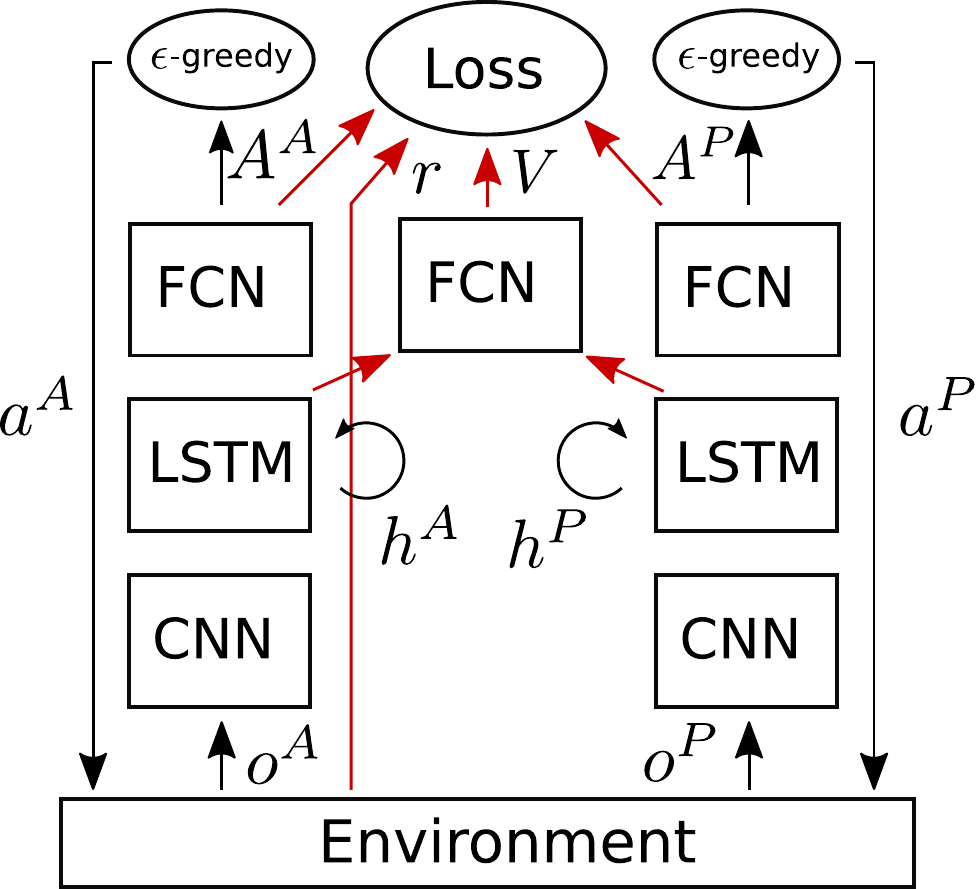}
	    \caption{\footnotesize MADDRQN}
		\label{fig:maddrqn}
	\end{subfigure}
	\caption{\footnotesize Information flow for the two models used in our experiments; red paths are only needed during training. The MADDRQN model (\subref{fig:maddrqn}) uses a centralized value-function with per-agent advantage functions. The centralized value function is only used during training. Superscripts $A$ and $P$ refer to the assistant and principal agents respectively. The MAIDRQN model (\subref{fig:maidrqn}) is used in experiments 1-3 and the MADDRQN model (\subref{fig:maddrqn}) is used in experiment 4 where it exhibits superior training characteristics for learning from pixels.}
	\label{fig:models}
\end{figure}

\section{The LILA Training Framework}
\label{sec:methods}

We now describe our training framework for producing an assisting agent that can learn a task interactively from a human user.
We define a task to be an instance of a cooperative partially observable Markov game as described in Section~\ref{sec:preliminaries}, with $N = 2$.
To enable the agent to solve such tasks, we train the agent, whom we call the ``assistant'' (superscript $A$), jointly with another agent, whom we call the ``principal'' (superscript $P$) on a variety of tasks. Critically, the principal's observation function informs it of the task.\footnote{Our tasks are similar to tasks in \citet{hadfield2016cir}, but with partially observable state and without access to the other agent's actions, which should better generalize to learning from humans in natural environment.}
The principal agent acts as a human surrogate which allows us to replace it with a human once the training is finished.
By informing the principal of the current task and withholding rewards and gradient updates until the end of each task, the agents are encouraged to emerge interactive learning behaviors in order to inform the assistant of the task and allow them to contribute to the joint reward.
We limit actions and observations to the physical environment, with the hope of emerging human-compatible behaviors.

In order to train the agents, we consider two different models. We first introduce a simple model that we find works well in tabular environments. Then, in order to scale our approach to pixel observations, we introduce a modification to the first model that we found was important in increasing the stability of learning.

\textbf{Multi-Agent Independent DRQN (MAIDRQN)}: 
The first model uses two deep recurrent $Q$-networks (DRQN)~\citep{hausknecht2015drq} that are each trained with Q-learning~\citep{watkins1989ldr}.
Let $Q_{\theta^i}(o^i_t, a^i_t, h^i_t)$ be the action-value function for agent $i$, which maps from the current action, observation, and history, $h^i_t$, to the expected discounted sum of future rewards.
The MAIDRQN method optimizes the following loss:\footnote{In our experiments we do not use a lagged ``target'' $Q$-network~\citep{mnih2013pad}, but we do stop gradients through the $Q$-network in $y_t$.}
\begin{align} \label{eqn:maidrqn_loss}
  \mathcal{L}^{\text{MAIDRQN}} &:= \frac{1}{N}\sum_{i,t}[y_t - Q_{\theta^i}(o^i_t,a^i_t,h^i_t)]^2 \\
  y_t &:= r_t + \gamma \max_{a^i_{t+1}}Q_{\theta^i}(o^i_{t+1}, a^i_{t+1}, h^i_{t+1}) \nonumber
\end{align}
The networks are trained simultaneously.
The model architecture is a recurrent neural network, depicted in Figure~\ref{fig:maidrqn}; see Section~\ref{sec:appendix_maidrqn} for details.
We use this model for experiments 1-3.

\begin{table*}[t]
\caption{\footnotesize Experimental configurations for our 4 experiments. Experiment 1 has two sub experiments, 1A and 1B. In 1B, the agents incur a penalty whenever the principals moves. The Observation Window column lists the radius of cells visible to each agent.}
\label{tbl:experimental_setups}
\begin{center}
\begin{small}
\begin{sc}
\begin{tabular}{lcccccc}
\toprule
\makecell{Exp.} & \makecell{Model} & \makecell{Principal\\Motion\\Penalty} & \makecell{Grid\\Shape} & \makecell{Num.\\Objects} & \makecell{Observations} & \makecell{Observation\\Window}\\
\midrule
1a & MAIDRQN & 0.0 & 5x5 & 10 & Binary Vectors & Full\\
1b & MAIDRQN & -0.4 & 5x5 & 10 & Binary Vectors & Full\\
2 & MAIDRQN & 0.0 & 5x5 & 10 & Binary Vectors & 1-Cell\\
3 & MAIDRQN & -0.1 & 3x1 ``L'' & 1 & Binary Vectors & 1-Cell\\
4 & MADDRQN & 0.0 & 5x5 & 10 & $64\times64\times3$ Pixels & $\sim$2-Cells\\
\bottomrule
\end{tabular}
\end{sc}
\end{small}
\end{center}
\end{table*}

\textbf{Multi-Agent Dueling DRQN (MADDRQN)}: With independent Q-Learning, as in MAIDRQN, the other agent's changing behavior and unknown actions make it difficult to estimate the Bellman target $y_t$ in Equation~\ref{eqn:maidrqn_loss}, which leads to instability in training. This model addresses part of the instability that is caused by unknown actions.

If $Q^*(o,a,h)$ is the optimal action-value function, then the optimal value function is $V^*(o,h) = \argmax_aQ^*(o,a,h)$, and the optimal advantage function is defined as $A^*(o,a,h) = Q^*(o,a,h) - V^*(o,h)$~\citep{wang2015dna}. The advantage function captures how inferior an action is to the optimal action in terms of the expected sum of discounted future rewards. This allows us to express $Q$ in a new form, $Q^*(o,a,h) = V^*(o,h) + A^*(o,a,h)$. We note that the value function is not needed when selecting actions: $\argmax_aQ^*(o,a,h) = \argmax_a(V^*(o,h) + A^*(o,a,h)) = \argmax_aA^*(o,a,h)$. We leverage this idea by making the following approximation to an optimal, centralized action-value function for multiple agents: 
\begin{align}\label{eqn:maddrqn_approx}
Q^*(\{o^i,a^i,h^i\}) &= V^*(\{o^i,h^i\}) + A^*(\{o^i,a^i,h^i\})\\
&\approx  V^*(\{o^i,h^i\}) + \sum_iA^{i*}(o^i, a^i, h^i), \nonumber
\end{align}
where $A^{i*}(o^i, a^i, h^i)$ is an advantage function for agent $i$ and $V^*(\{o^i,h^i\})$ is a joint value function.\footnote{
The approximation is due to the substitution of $\sum_iA^{i*}(o^i,a^i,h^i)$ for $A^*(\{o^i,a^i,h^i\})$ in Equation~\ref{eqn:maddrqn_approx}, which implies that the agents' current actions have independent effects on expected future rewards, and is not true in general. Nevertheless, it is a useful approximation.
}

The training loss for this model is:
\begin{align}\label{eqn:maddrqn_loss}
  \mathcal{L}^{\text{MADDRQN}} &:= \sum_{t}[y_t - Q_{\{\theta^i\},\phi}(\{o^i_t,a^i_t,h^i_t\})]^2\\
  y_t &:= r_t + \gamma\max_{\{a^i_{t+1}\}}Q_{\{\theta^i\},\phi}(\{o^i_{t+1},a^i_{t+1},h^i_{t+1}\}) \nonumber
\end{align}
where
\begin{equation}\label{eqn:maddrqn}
Q_{\{\theta^i\},\phi}(\{o^i_t,a^i_t,h^i_t\}) := V_\phi(\{o^i_t,h^i_t\}) + \sum_iA_{\theta^i}(o^i_t,a^i_t,h^i_t).
\end{equation}
Once trained, each agent selects their actions according to their advantage function $A_{\theta^i}$,
\begin{equation}
a^i_t = \argmax_aA_{\theta^i}(o^i_t, a, h^i_t),
\end{equation}
as opposed to the Q-functions $Q_{\theta^i}$ in the case of MAIDDRQN.

In the loss for the MAIDRQN model, Equation~\ref{eqn:maidrqn_loss}, there is a squared error term for each $Q_{\theta^i}$ which depends on the joint reward $r$. This means that, in addition to estimating the immediate reward due their own actions, each $Q_{\theta^i}$ must estimate the immediate reward due to the actions of the other agent, without access to their actions or observations. By using a joint action value function and decomposing it into advantage functions and a value function, each $A^i$ can ignore the immediate reward due to the other agent, simplifying the optimization.

We refer to this model as a multi-agent dueling deep recurrent $Q$-network (MADDRQN), in reference to the single agent dueling network of~\citet{wang2015dna}. The MADDRQN model, which adds a fully connected network for the shared value function, is depicted in Figure~\ref{fig:maddrqn}; See Section~\ref{sec:appendix_maddrqn} for details. The MADDRQN model is used in experiment 4.

\textbf{Training Procedure}:
We use a standard episodic training procedure, with the task changing on each episode. 
The training procedure for the MADDRQN and MAIDRQN models differ only in the loss function.
Here, we describe the training procedure with reference to the MADDRQN model.
We assume access to a subset of tasks, $\DTrain$, from a task domain, $\D = \{..., \task_j, ...\}$.
First, we initialize the parameters $\theta^P$, $\theta^A$, and $\phi$. Then, the following procedure is repeated until convergence. A batch of tasks are uniformly sampled from $\DTrain$. For each task $\task_b$ in the batch, a trajectory, $\tau_b=(o^P_0,o^A_0,a^P_0,a^A_0,r_0...,o^P_H,o^A_H,a^P_H,a^A_H,r_H)$, is collected by playing out an episode in an environment configured to $\task_b$, with actions chosen $\epsilon$-greedy according to $A_{\theta^P}$ and $A_{\theta^A}$. The hidden states for the recurrent LSTM cells are reset to $\mathbf{0}$ at the start of each episode. The loss for each trajectory is calculated using Equations~\ref{eqn:maddrqn_loss} and \ref{eqn:maddrqn}. Finally, a gradient step is taken with respect to $\theta^P$, $\theta^A$, and $\phi$ on the sum of the episode losses.

\begin{table*}[t]
\caption{\footnotesize Results for Experiments 1A and 1B. Experiment 1B includes a motion penalty for the principal's motion. In both experiments, MAIDRQN outperforms the principal acting alone, demonstrating that the assistant learns from and assists the principal. All performance increases are significant ($\text{confidence}>99\%$), except for FeedFwd-A and Solo-P in Experiment 1A, which are statistically equivalent.}
\label{tbl:exp1}
\begin{center}
\begin{small}
\begin{sc}
\begin{tabular}{lcc|cccc|cc}
\toprule
& & \multicolumn{3}{c}{Experiment 1A} & & \multicolumn{3}{c}{Experiment 1B}\\
\midrule
\makecell{Method\\Name} & \hspace{0.08in} & \makecell{Joint\\Reward} & \makecell{Reward\\due to P} & \makecell{Reward\\due to A} & \hspace{0.08in} & \makecell{Joint\\Reward} & \makecell{Reward\\due to P} & \makecell{Reward\\due to A}\\
\midrule
Oracle-A & & 4.9 $\pm$ 0.2 & 2.5 $\pm$ 0.1 & 2.4 $\pm$ 0.1 & & 4.0 $\pm$ 0.1 & 0.0 $\pm$ 0.0 & 4.0 $\pm$ 0.1\\
\midrule
MAIDRQN & & \textbf{4.6 $\pm$ 0.2} & 3.3 $\pm$ 0.2 & 1.3 $\pm$ 0.3 & & \textbf{3.6 $\pm$ 0.1} & 0.4 $\pm$ 0.1 & 3.2 $\pm$ 0.1\\
FeedFwd-A & & 4.1 $\pm$ 0.1 & 4.1 $\pm$ 0.1 & 0.0 $\pm$ 0.0\footnotemark & & 2.0 $\pm$ 0.4 & 0.7 $\pm$ 0.3 & 1.3 $\pm$ 0.6\\
Solo-P & & 4.0 $\pm$ 0.1 & 4.0 $\pm$ 0.1 & N/A & & 1.2 $\pm$ 0.1 & 1.2 $\pm$ 0.1 & N/A\\
\bottomrule
\end{tabular}
\end{sc}
\end{small}
\end{center}
\end{table*}

\begin{figure*}[t!]
  \begin{minipage}[t!]{0.49\textwidth}
  \centering
  \includegraphics[width=0.90\textwidth]{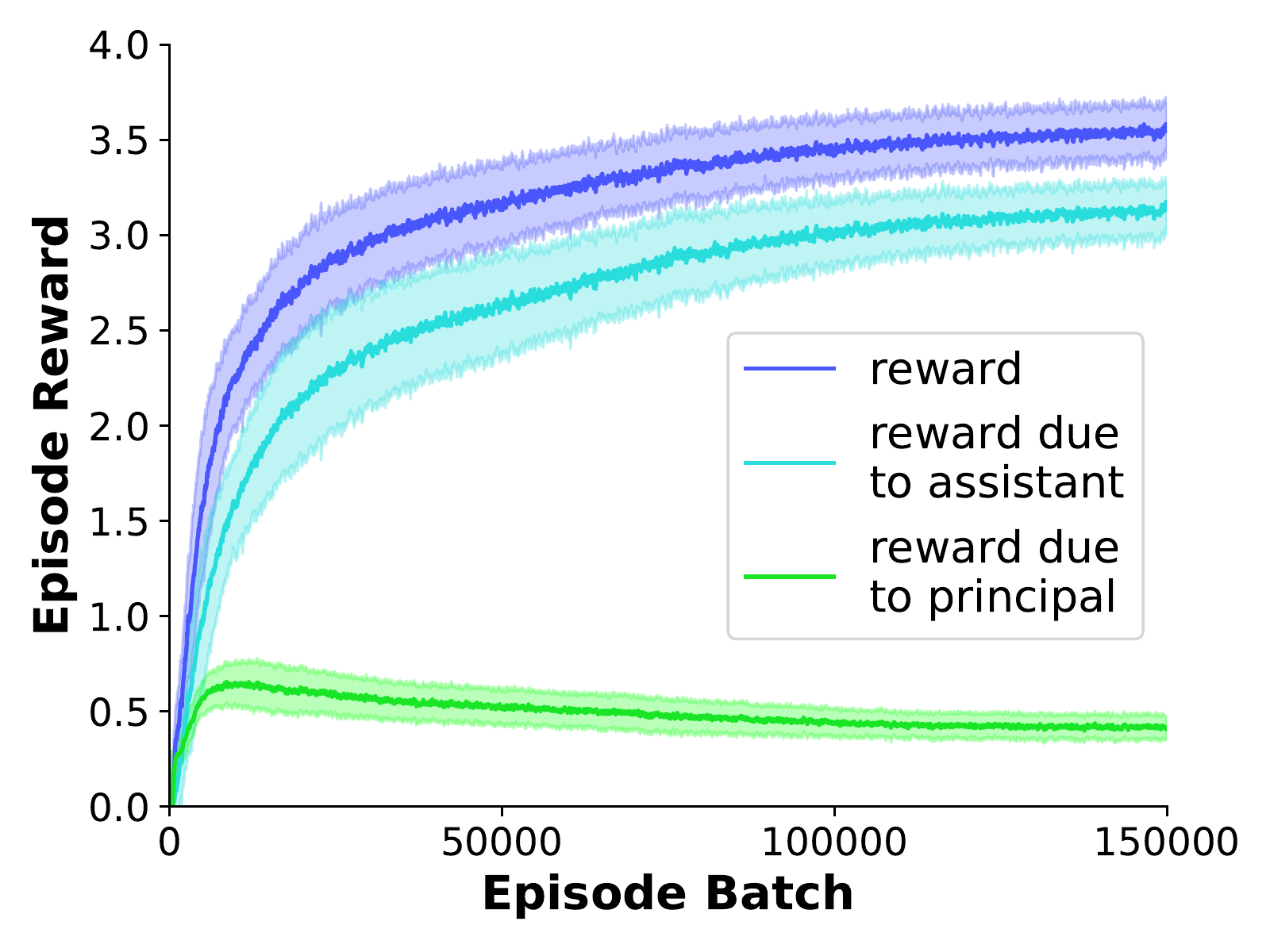}
  \caption{\footnotesize
  Training curve for Experiment 1B.
  Error bars are 1 standard deviation. 
  At the end of training, nearly all of the joint reward in an episode is due to the assistant's actions, indicating that the trained assistant can learn the task and then complete it independently.
  }
  \label{fig:episode_rewards}
  \end{minipage}
  ~
  \begin{minipage}[t!]{0.49\textwidth}
  \centering
  \begin{subfigure}[b]{0.48\textwidth}
    \includegraphics[width=1.6in]{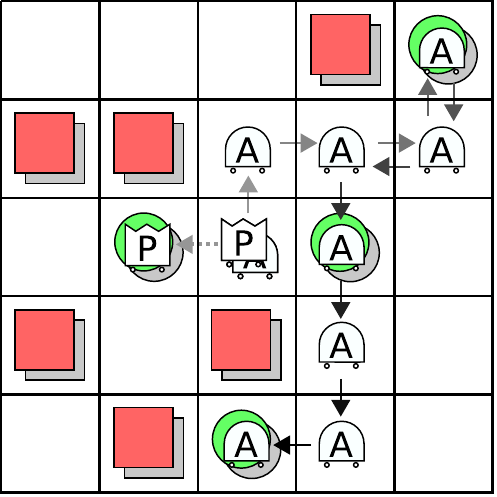}
    \caption{\footnotesize The assistant learns from a single principal movement.}
  \end{subfigure}
  ~
  \begin{subfigure}[b]{0.48\textwidth}
    \includegraphics[width=1.6in]{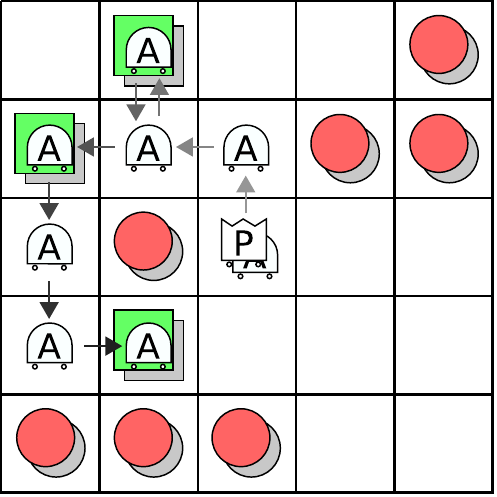}
    \caption{\footnotesize The assistant learns from a lack of principal movement.}
  \end{subfigure}
  \caption{\footnotesize 
  Episode traces of trained agents on test tasks from Experiment 1B. Both agents begin in the center square and cooperatively collect as many instances of the target shape as possible. The target shape is shown in green. The principal agent $P$ observes the target shape, but the assistant agent $A$ does not and must learn from the principal's movement or lack of movement.
  The assistant rapidly learns the target shape from the principal and collects all instances.
  }
  \label{fig:exp1_n04rm_traces}
  \end{minipage}
\end{figure*}

\section{Experimental Results}
\label{sec:results}

We designed a series of experiments in order to study how different interactive learning behaviors may emerge, to test whether our method can scale to pixel observations, and to evaluate the ability for the agents to transfer to a setting with a human user.

We conducted four experiments on grid-world environments, where the goal was to cooperatively collect all objects from one of two object classes.
Two agents, the principal and the assistant, act simultaneously and may move in one of the four cardinal directions or may choose not to move, giving five possible actions per agent.

Within an experiment, tasks vary by the placement of objects, and by the class of objects to be collected, which we call the ``target class''.
The target class is supplied to the principal as a two dimensional, one-hot vector.
If either agent enters a cell containing an object, the object disappears and both agents receive a reward: $+1$ for objects of the target class and $-1$ otherwise. 
Each episode consisted of a single task and lasted for 10 time-steps.
Table~\ref{tbl:experimental_setups} gives the setup for each experiment.

We collected 10 training runs per experiment, and we report the aggregated performance of the 10 trained agent pairs on 100 test tasks not seen during training. The training batch size was 100 episodes and the models were trained for 150,000 gradient steps (Experiments 1-3) or 40,000 gradient steps (Experiment 4).
\footnotetext{The FeedForward assistant moves 80\% of the time, but it never collects an object.} Videos for all experiments, as well as an interactive game, are available on the paper website.\footnote{\url{https://interactive-learning.github.io}}

\textbf{Experiment 1 A\&B -- Learning and Assisting:} 
In this experiment we explore if the assistant can be trained to learn and assist the principal.
Table~\ref{tbl:exp1} shows the experimental results without and with a penalty for motion of the principal (Experiments 1A and 1B respectively).
Figures~\ref{fig:episode_rewards}~and~\ref{fig:exp1_n04rm_traces} show the learning curve and trajectory traces for trained agents in Experiment 1B.

The joint reward of our approach (MAIDRQN) exceeds that of a principal trained to act alone (Solo-P), and approaches the optimal setting where the assistant also observes the target class (Oracle-A). Further, we see that the reward due to the assistant is positive, and even exceeds the reward due to the principal when the motion penalty is present (Experiment 1B). This demonstrates that the assistant learns the task from the principal and assists the principal. Our approach also outperforms an ablation in which the assistant's LSTM is replaced with a feed forward network (FeedFwd-A), highlighting the importance of memory.

\begin{figure*}[t]
  \centering
  \begin{minipage}[t]{0.49\textwidth}
  \centering
  \begin{subfigure}[t]{0.48\textwidth}
    \includegraphics[width=1.6in]{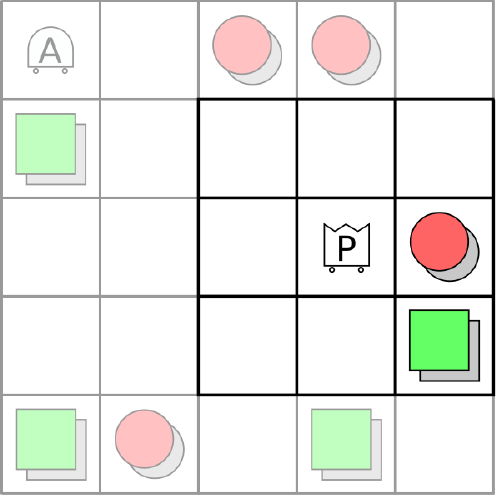}
    \caption{\footnotesize Principal View}
  \end{subfigure}
  ~
  \begin{subfigure}[t]{0.48\textwidth}
    \includegraphics[width=1.6in]{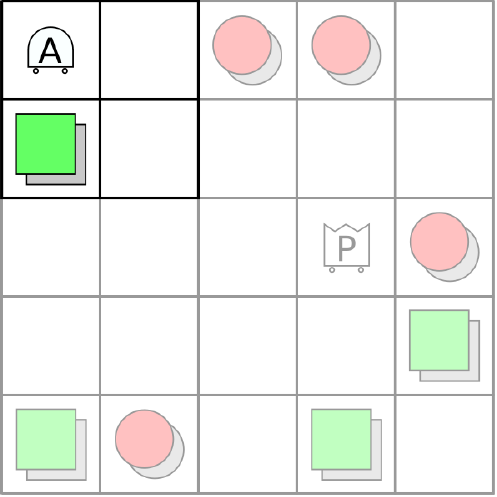}
    \caption{\footnotesize Assistant View}
  \end{subfigure}
  \caption{\footnotesize
  Visualization of the 1-cell observation window used in experiments 2 and 3.  Cell contents outside of an agent's window are hidden from that agent.
  }
  \label{fig:limited_view}
  \end{minipage}
  ~
  \begin{minipage}[t]{0.49\textwidth}
  \centering
  \begin{subfigure}[t]{0.48\textwidth}
	\includegraphics[width=1.6in]{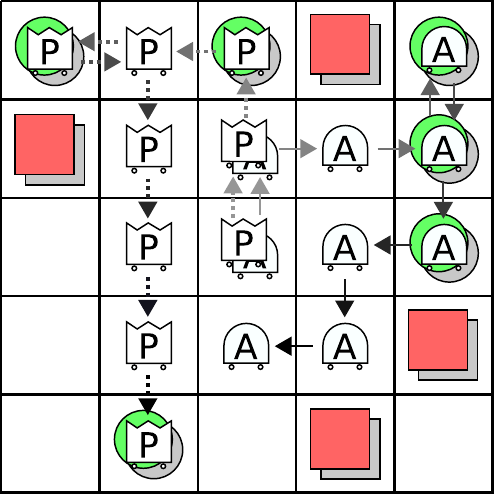}
	\caption{\footnotesize 2 step info. seek}
	\label{fig:seek_info_seek_info}
  \end{subfigure}
  ~
  \begin{subfigure}[t]{0.48\textwidth}
	\includegraphics[width=1.6in]{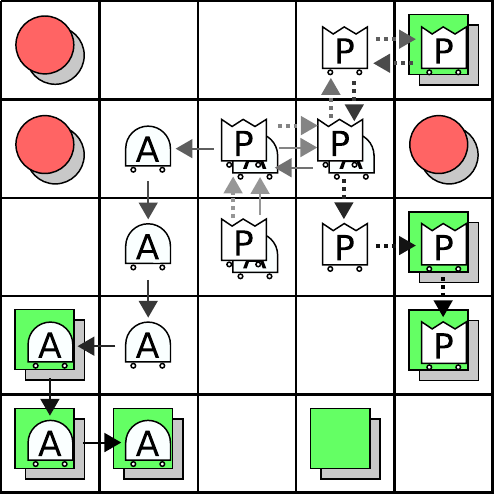}
	\caption{\footnotesize 3 step info. seek}
	\label{fig:seek_info_multistep_seek_info}
  \end{subfigure}
  \caption{\footnotesize
  Episode traces of trained agents on test tasks from Experiment 2. Both agents begin in the center square and cooperatively collect as many instances of the target shape as possible. The target shape is shown in green. The principal agent $P$ observes the target shape, but the assistant agent $A$ does not and must learn from the principal's movement
  With restricted observations, the assistant moves with the principal until it observes a disambiguating action, and then proceeds to collect the target shape on its own.
  }
  \label{fig:seek_info}
  \end{minipage}
\end{figure*}

\textbf{Experiment 2 -- Active Information Gathering:}
In this experiment we explore if, in the presence of additional partial observability, the assistant will take actions to actively seek out information.
This experiment restricts the view of each agent to a 1-cell window and only places objects around the exterior of the grid, requiring the assistant to move with the principal and observe its behavior, see Figure~\ref{fig:limited_view}.
Figure~\ref{fig:seek_info} shows trajectory traces for two test tasks.
The average joint reward, reward due to the principal, and reward due to the assistant are $4.7 \pm 0.2$, $2.8 \pm 0.2$, and $1.9 \pm 0.1$ respectively.
This shows that our training framework can produce information seeking behaviors.

\begin{figure*}[t]
  \centering
  \begin{subfigure}[t]{0.98\textwidth}
    \centering
    \includegraphics[width=0.8\linewidth]{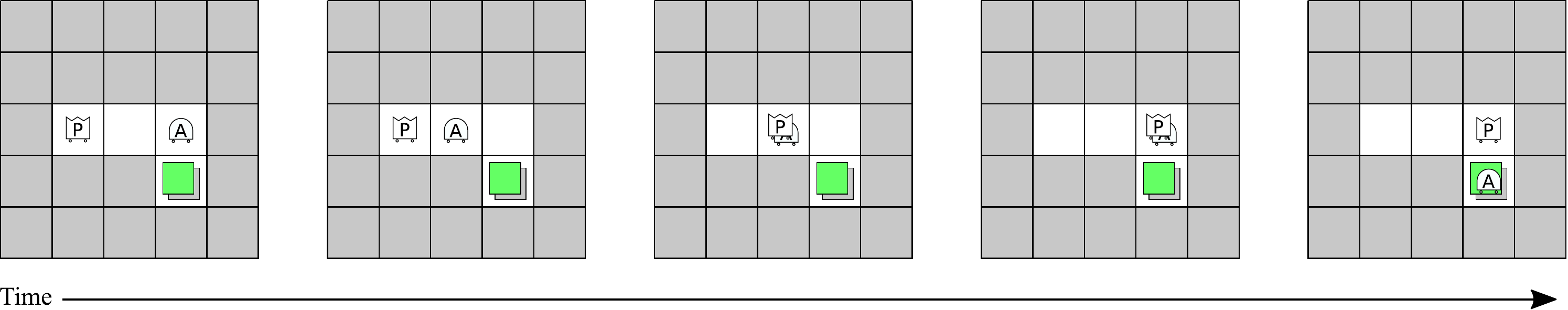}
    \caption{\footnotesize The square should be collected (green), but the assistant does not observe this (grey under green).}
  	\label{fig:question_ug}
  \end{subfigure}
  \par\medskip
  \begin{subfigure}[t]{0.98\textwidth}
    \centering
	\includegraphics[width=0.8\linewidth]{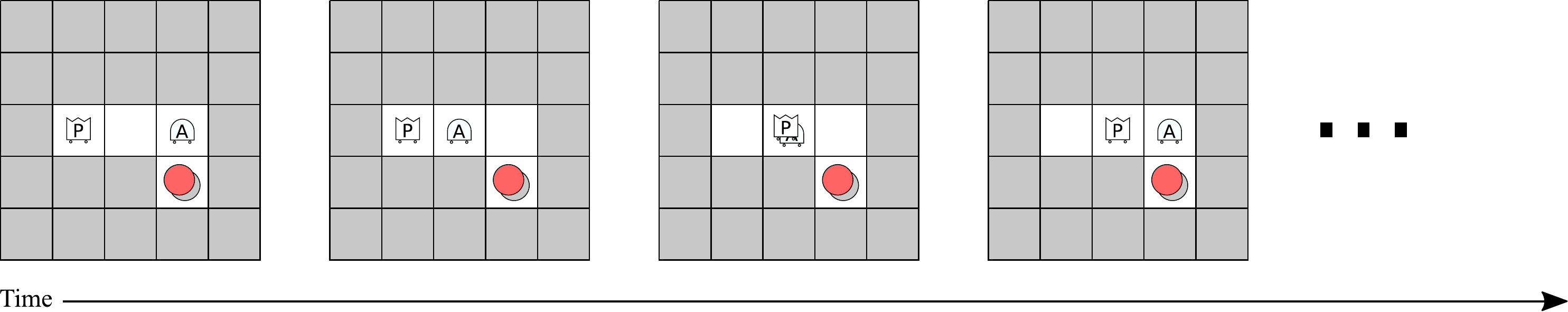}
	\caption{\footnotesize The circle should not be collected (red), but the assistant does not observe this (grey under red)}
  	\label{fig:question_ub}
  \end{subfigure}
  \caption{\footnotesize
    Episode roll-outs for trained agents from Experiment 3. When the assistant is uncertain of an object, it requests information from the principal by moving into its visual field and observing the response.
  }
  \label{fig:question}
\end{figure*}

\vspace{0.1cm}
\textbf{Experiment 3 -- Interactive Questioning and Answering:}
In this experiment we explore if there is a setting where explicit questioning and answering can emerge.
On 50\% of the tasks, the assistant is allowed to observe the target class.
This adds uncertainty for the principal, and discourages it from proactively informing the assistant.
Figure~\ref{fig:question} shows the first several states of tasks in which the assistant does not observe the target class.\footnote{The test and training sets are the same in Experiment 3, since there are only 8 possible tasks}

The emerged behavior is for the assistant to move into the visual field of the principal, effectively asking the question, then the principal moves until it sees the object, and finally answers the question by moving one step closer only if the object should be collected.
The average joint reward, reward due to the principal, and reward due to the assistant are $0.4 \pm 0.1$, $-0.1 \pm 0.1$, and $0.5 \pm 0.1$ respectively. 
This demonstrates that our framework can emerge question-answering, interactive behaviors.

\begin{figure}[t]
    \centering
    \begin{subfigure}[t]{0.23\textwidth}
	    \centering
	    \includegraphics[height=1.25in]{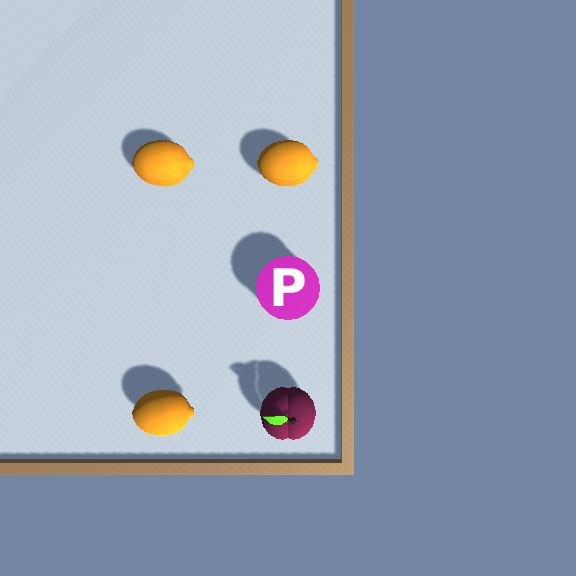}
        \caption{\footnotesize Principal}
	    \label{fig:exp4_observation_principal}
    \end{subfigure}
    \begin{subfigure}[t]{0.23\textwidth}
	    \centering
	    \includegraphics[height=1.25in]{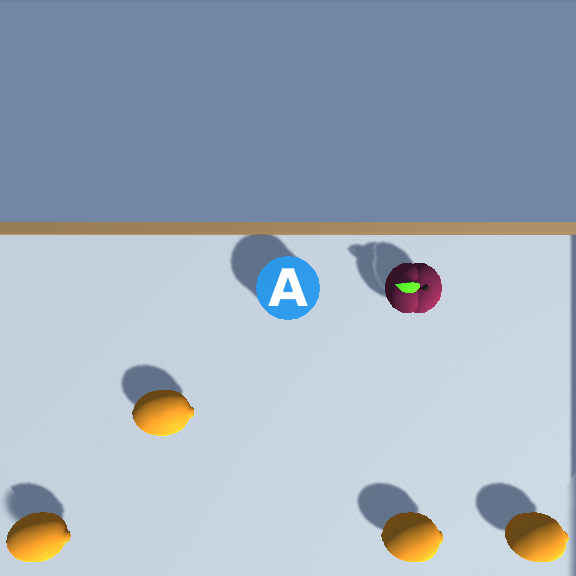}
        \caption{\footnotesize Assistant}
	    \label{fig:exp4_observation_assistant}
    \end{subfigure}
	\caption{\footnotesize 
	Example observations for experiment 4. 
	The principal's observation also includes a 2 dimensional one-hot vector indicating the fruit to collect, plums in this case. 
	These are the 7th observations from the human-agent trajectory in Figure~\ref{fig:exp4_traces_human}.}
	\label{fig:exp4_observations}
\end{figure}

\textbf{Experiment 4 -- Learning from and Assisting a Human Principal with Pixel Observations:}
In this final experiment we explore if our training framework can extend to pixel observations and whether the trained assistant can learn from a human principal.
Figure~\ref{fig:exp4_observations} shows examples of the pixel observations.
Ten participants, who were not familiar with this research, were paired with the 10 trained assistants, and each played 20 games with the assistant and 20 games without the assistant. Participants were randomly assigned which setting to play first.
Figure~\ref{fig:exp4_traces} shows trajectory traces on test tasks at several points during training and with a human principal after training.

Unlike the previous experiments, stability was a challenge in this problem setting; most training runs of MAIDRQN became unstable and dropped below 0.1 joint reward before the end of training.
Hence, we chose to use the MADDRQN model because we found it to be more stable than MAIDRQN.
The failure rate was 64\% vs 75\% for each method respectively, and the mean failure time was 5.6 hours vs 9.7 hours ($\text{confidence}>99\%$), which saved training time and was a practical benefit.

Table~\ref{tbl:exp4} shows the experimental results.
The participants scored significantly higher with the assistant than without  ($\text{confidence}>99\%$). This demonstrates that our framework can produce agents that can learn from humans.

\begin{table}[t]
\caption{\footnotesize Results for Experiment 4. Trained assistants learned from human principals and significantly increased their scores (Human\&Agent) over the humans acting alone (Human), demonstrating the potential for our training framework to produce agents that can learn from and assist humans.\protect\footnotemark}
\label{tbl:exp4}
\begin{center}
\begin{small}
\begin{sc}
\begin{tabular}{lc|cc}
\toprule
\makecell{Players} & \makecell{Joint\\Reward} & \makecell{Reward\\due to P} & \makecell{Reward\\due to A}\\
\midrule
Agent\&Agent    & 4.6 $\pm$ 0.2 & 2.6 $\pm$ 0.2 & 2.0 $\pm$ 0.2\\
Human\&Agent    & 4.2 $\pm$ 0.4 & 2.9 $\pm$ 0.3 & 1.3 $\pm$ 0.5\\
Agent           & 3.9 $\pm$ 0.1 & 3.9 $\pm$ 0.1 & N/A\\
Human           & 3.8 $\pm$ 0.3 & 3.8 $\pm$ 0.3 & N/A\\
\bottomrule
\end{tabular}
\end{sc}
\end{small}
\end{center}
\end{table}
\footnotetext{Significance is based on a t-test of the participants' change in score, which is more significant than the table's standard deviations would suggest ($\text{confidence} > 99\%$).}

While inclusion of the assistant increases the human's score, it is still less than the score when the assistant acts with the principal agent with which it was trained. What is the cause of this gap? To answer this question, we identified that 12\% of the time the assistant incorrectly infers which object to collect (episodes where the assistant always collects the wrong object). If we exclude these episodes, we obtain the performance when the assistant has correctly inferred the task, but must still coordinate with the human principal. Humans with ``correct'' assistants achieve rewards ($4.7\pm-0.3$, $2.8\pm0.3$, $1.9\pm0.2$) that are statistically equivalent to Agent\&Agent and statistically superior to Human\&Agent in Table~\ref{tbl:exp4}. This means that the assistants coordinate equivalently with human principals and artificial principals, but can experience problems inferring the task from human principals, resulting in the observed drop in score. This is an example of co-adaptation to communicating with the principal agent during training. The next section suggests an approach to address such co-adaptation.

\section{Summary and Future Work}
\label{sec:summary_and_future_work}

We introduced the LILA training framework, which trains an assistant to learn interactively from a knowledgeable principal through only physical actions and observations in the environment.
LILA produces the assistant by jointly training it with a principal, who is made aware of the task through its observations, on a variety of tasks, and restricting the observation and action spaces to the physical environment.
We further introduced the MADDRQN algorithm, in which the agents have individual advantage functions but share a value function during training. 
MADDRQN showed improved stability over MAIDRQN, which was a practical benefit in the experiments.
The experiments demonstrate that, depending on the environment, LILA emerges behaviors such as demonstrations, partial demonstrations, information seeking, and question answering.
Experiment 4 demonstrated that LILA scales to environments with pixel observations, and, crucially, that LILA is able to produce agents that can learn from and assist humans.

A possible future extension involves training with populations of agents.
In our experiments, the agents sometimes emerged overly co-adapted behaviors. For example, in Experiment 2, the agents tend to always move in the same direction in the first time step, but the direction varies by the training run.
This makes agents paired across runs less compatible, and less likely to generalize to human principals.
We believe that training an assistant across populations of agents will reduce such co-adapted behaviors.
Finally, LILA's emergence of behaviors means that the trained assistant can only learn from behaviors that emerged during training. Further research should seek to minimize these limitations, perhaps through advances in online meta-learning~\cite{online_meta_learning}.

\subsubsection{Acknowledgments}
The authors are especially grateful to Alonso Martinez for designing and iterating on the Unity environment used in Experiment 4.

{
\bibliography{AAAI-WoodwardM.5768}}
\bibliographystyle{aaai}

\appendix

\clearpage
\section{Appendix}

\subsection{Model Architectures}

\subsubsection{Multi-Agent Independent DRQN (MAIDRQN)}
\label{sec:appendix_maidrqn}

The MAIDRQN model makes use of independent DRQN models, see figure~\ref{fig:maidrqn}.
There are independent parallel paths for each agent, giving $Q^P$ and $Q^A$.
Each path consists of a convolutional neural network (CNN)~\citep{lecun1995cni}, followed by a long short-term memory (LSTM)~\citep{hochreiter1997lst}, followed by a fully-connected layer.
The CNNs consist of a 2-layer network with 10, 3x3, stride 1 filters per layer and rectified linear unit (ReLU) activations. The LSTMs have 50 hidden units. The fully-connected network layer outputs are 5 dimensional, representing $Q$ values for the 5 actions in the environment.

\subsubsection{Multi-Agent Dueling DRQN (MADDRQN)}
\label{sec:appendix_maddrqn}

Up to the final layer, the MADDRQN model is similarly structured to the MAIDRQN model, with the following modifications. The CNN filter weights are shared between the agents. The first CNN layer has 16, 8x8 filters with stride 4, and the second layer has 32, 8x8 filters with stride 2, both with ReLU activations. A 256-unit fully-connected layer with ReLU activations is inserted between the CNN and the LSTM, also with weights shared between the agents. Information about the current task is represented as a 1-hot vector and concatenated to the output of the fully-connected layer before feeding it to the principal's LSTM. The outputs for each LSTM are fed to separate fully-connected layers with 5 outputs each, and the softmax across the 5 outputs is subtracted from each output, producing an advantage functions, $A^i$, for each agent. In a separate path, the outputs from the two LSTMs are concatenated together and fed to a fully-connected layer with a single output, representing the shared joint value function $V$. The MADDRQN model is depicted in Figure~\ref{fig:maddrqn}

\subsubsection{Multi-Agent Independent DRQN (MAIDRQN) (Experiment 4)}
\label{sec:appendix_maidrqn_exp4}

In Experiment 4, we conducted a comparison between MADDRQN and MAIDRQN. For a fair comparison, the MAIDRQN model in Experiment 4 is the MADDRQN model without the value function, and without the softmax subtracted from the fully-connected output layers, giving $Q^P$ and $Q^A$. All other aspects, including the fully-connected layer after the CNN layers and the sharing of weights are identical to MADDRQN.

\subsection{Experimental Details}
\label{sec:appendix_experimental_details}

All experiments use $\epsilon$-greedy exploration during training ($\epsilon = 0.05$), and argmax when evaluating. The Adam optimizer with default parameters was used to train the models~\cite{kingma2015ams}.

All experiments operate on a grid. On every step, each agent can take one of five actions: move left, move right, move up, move down, or stay in place. Actions are taken in parallel, and the subsequent observations reflect the effect of both agents' actions. If either agent moves onto a cell containing an object, then the object is removed and considered ``collected'', with resulting joint reward defined by $R$. The agents may occupy the same cell in the shape environment of Experiments 1-3, but collide with each other in the fruit environment of Experiment 4; if, in the fruit environment, the agents attempt to move onto the same cell, the principal gets to move to the cell and the assistant remains at its current location.

In experiments using the shape environment, the agents observe the world as a grid of binary vectors, one binary vector for each cell. Some fields in the binary vector are zeroed out depending on the agent and the experiment. Each binary vector contains: 1 bit indicating if the cell is visible to the agent, 1 bit indicating the presence of the principal, 1 bit indicating the presence of the assistant, 1 bit indicating the presence of an object, 2 bits indicating the class of the object in one-hot form, 2 bits indicating whether the object should be collected in one-hot form. All bits are zeroed if the cell is not visible to the agent, and all object bits are zeroed if no object is present in the cell.

In Experiment 1, the bits indicating whether the object should be collected are set to zero in the assistant's binary vectors.

In Experiment 2, the bits indicating whether the object should be collected are set to zero in the assistant's binary vectors. Further, for both agents, if a cell is more than 1 cell away from the agent, then all bits in the binary vector for that cell are set to zero.

In Experiment 3, on half of the tasks, the bits in indicating whether the object should be collected are set to zero in the assistant's binary vectors. Further, for both agents, if a cell is more than 1 cell away from the agent, then all bits in the binary vector for that cell are set to zero.

In Experiment 4, using the fruit environment, the agents observe the world as a 64x64x3 color image centered on the respective agent, see Figure~\ref{fig:exp4_observations}. The camera view includes the entire world when the agent is at the center of the grid, but becomes partially observable as the agent moves away from the center. The principal agent receives the class to collect, in one-hot form, as an additional observation. The class of objects to collect is concatenated to the principal's LSTM input. In Experiment 4, the assistant never observes the class of objects to collect.

\subsection{Training Algorithms}
\label{sec:appendix_algorithms}

\begin{algorithm}[t!]
   \caption{\footnotesize Training Procedure using the MAIDRQN model}
   \label{alg:maidrqn}
\begin{algorithmic}[1]
   \REQUIRE $\DTrain$: a set of training tasks
   \REQUIRE Two recurrent artificial neural networks that output action-value functions $Q_{\theta^P}$ and $Q_{\theta^A}$
   \REQUIRE An environment interface $Env$ with methods $init()$ and $step()$
   \ENSURE $\theta^P$ and $\theta^A$
   \STATE Initialize $\theta^P$ and $\theta^A$
   \WHILE{not done}
     \STATE Uniformly sample a batch of tasks from $\DTrain$, let $\task_b$ be a task in the batch.
     \FORALL{$\task_b$}
       \STATE Reset the memory for the recurrent neural networks to $\mathbf{0}$
       \STATE $o^P,o^A = Env.init(\task_b)$
       \STATE $\tau_b = [o^P,o^A]$
       \REPEAT
         \STATE Choose actions $a^P$ and $a^A$ using $\epsilon$-greedy exploration on $Q_{\theta^P}(o^P)$ and $Q_{\Theta^A}(o^A)$
         \STATE $r,o^P,o^A = Env.step(a^P,a^A)$
         \STATE $\tau_b.append(a^P, a^A, r, o^P, o^A)$
       \UNTIL{end of episode}
       \STATE Calculate $\mathcal{L}_b^{\text{MAIDRQN}}$ using Equation~\ref{eqn:maidrqn_loss} with $o^i_t$, $a^i_t$, and $r_t$ from $\tau_b$
     \ENDFOR
     \STATE Take a gradient step on $\sum_b\mathcal{L}_b^{\text{MAIDRQN}}$ w.r.t. $\theta^P$ and $\theta^A$
   \ENDWHILE
\end{algorithmic}
\end{algorithm}

\begin{algorithm}[t!]
   \caption{\footnotesize Training Procedure using the MADDRQN model}
   \label{alg:maddrqn}
\begin{algorithmic}[1]
   \REQUIRE $\DTrain$: a set of training tasks
   \REQUIRE A recurrent artificial neural network that outputs advantage functions $A_{\theta^P}$ and $A_{\theta^A}$, and value function $V_\phi$
   \REQUIRE An environment interface $Env$ with methods $init()$ and $step()$
   \ENSURE $\theta^P$, $\theta^A$, and $\phi$
   \STATE Initialize $\theta^P$, $\theta^A$, and $\phi$
   \WHILE{not done}
     \STATE Uniformly sample a batch of tasks from $\DTrain$, let $\task_b$ be a task in the batch
     \FORALL{$\task_b$}
       \STATE Reset the memory for the recurrent neural networks to $\mathbf{0}$
       \STATE $o^P,o^A = Env.init(\task_b)$
       \STATE $\tau_b = [o^P,o^A]$
       \REPEAT
         \STATE Choose actions $a^P$ and $a^A$ using $\epsilon$-greedy exploration on $A_{\theta^P}(o^P)$ and $A_{\theta^A}(o^A)$
         \STATE $r,o^P,o^A = Env.step(a^P,a^A)$
         \STATE $\tau_b.append(a^P, a^A, r, o^P, o^A)$
       \UNTIL{end of episode}
       \STATE Calculate $\mathcal{L}_b^{\text{MADDRQN}}$ using Equations~\ref{eqn:maddrqn_loss} and \ref{eqn:maddrqn} with $o^i_t$, $a^i_t$, and $r_t$ from $\tau_b$
     \ENDFOR
     \STATE Take a gradient step on $\sum_b\mathcal{L}_b^{\text{MADDRQN}}$ w.r.t. $\theta^P$, $\theta^A$, and $\phi$
   \ENDWHILE
\end{algorithmic}
\end{algorithm}

Algorithm~\ref{alg:maidrqn} describes the training procedure for training with the Multi-Agent Independent DRQN (MAIDRQN) model, used in experiments 1-3. Algorithm~\ref{alg:maddrqn} describes the training procedure for training with our Multi-Agent Dueling DRQN (MADDRQN) model, used in experiment 4.

\subsection{Experiment 3 - Rollouts}
\label{sec:appendix_exp3_rollouts}

\begin{figure*}
  \centering
  \begin{subfigure}[t]{0.9\textwidth}
	\includegraphics[width=1.0\textwidth]{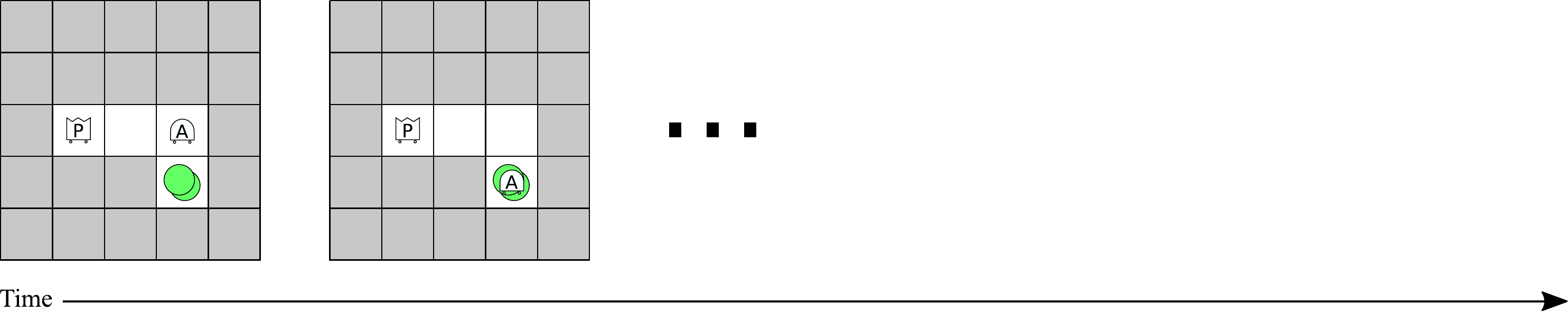}
	\caption{\footnotesize The assistant immediately consumes the good object when it observes object goodness}
	\label{fig:appendix_question_kg}
  \end{subfigure}
  \par\medskip
  \begin{subfigure}[t]{0.9\textwidth}
	\includegraphics[width=1.0\textwidth]{question_ugs}
	\caption{\footnotesize The assistant asks the principal whether the object is good when it does not observe object goodness.}
  	\label{fig:appendix_question_ug}
  \end{subfigure}
  \par\medskip
  \begin{subfigure}[t]{0.9\textwidth}
	\includegraphics[width=1.0\textwidth]{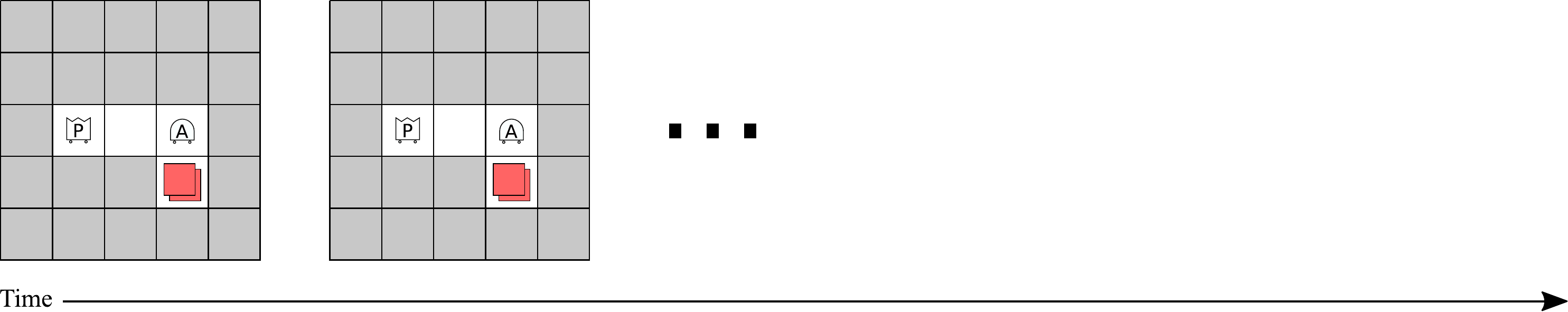}	
	\caption{\footnotesize The assistant ignores the bad object when it observes object goodness}
  	\label{fig:appendix_question_kb}
  \end{subfigure}
  \par\medskip
  \begin{subfigure}[t]{0.9\textwidth}
	\includegraphics[width=1.0\textwidth]{question_ubc}
	\caption{\footnotesize The assistant asks the principal whether the object is good when it does not observe object goodness.}
  	\label{fig:appendix_question_ub}
  \end{subfigure}
  \caption{\footnotesize 
	The assistant has learned how and when to ask questions to the principal. The assistant randomly observes object goodness dependent on the episode. When asking a question, as shown in cases (\subref{fig:appendix_question_ug}) and (\subref{fig:appendix_question_ub}), the assistant moves into the visual field of the principal. Next, the principal moves to the right to observe the object and moves one step closer if the object is good or does not move if the object is bad. The assistant understands this answer and proceed correctly.
  }
  \label{fig:appendix_question}
\end{figure*}

Figure~\ref{sec:appendix_exp3_rollouts} shows rollouts for emerged behaviors in 4 of the 8 tasks.

In the case where the assistant observes object goodness and the object is bad, Figure~\ref{fig:appendix_question_kb}, the assistant correctly does not collect the bad object.
However, it does move into the principal's visual field later in the episode. 
A late appearance in the principal's visual field must not be a well-formed question since the principal does not move to answer. 
The emergence of time-step-dependent behaviors such as this might be avoided by starting agents at different random steps in the episode.

\subsection{Experiment 4 - Human Instructions}

The following instructions were provided to the human in the Experiment 4.
\vspace{0.3cm}

Your goal in each episode is to harvest as many lemons or plums as you can. You will be shown which fruit to harvest, lemons or plums, in a separate window. You will receive +1 for each correct fruit harvested and -1 for each incorrect fruit harvested. An episode lasts for 10 steps, i.e. 10 action choices. 

You will play 20 episodes by yourself, controlling the principal agent, “P”. You will also play 20 episodes with an assistant agent, “A”. The assistant is a trained neural network. The assistant does not know which fruit to harvest at the start of the episode, it must learn the correct fruit from you. When harvesting with the assistant, you will receive +1 and -1 for each correct and incorrect fruit harvested, regardless of which agent harvested the fruit.

You will begin with 10 practice episodes for each agent setting (with the assistant and without).

\noindent
Controls: 
\begin{itemize}
\item \textless{}Enter\textgreater{}: starts an episode, the target fruit will be displayed
\item \textless{}Space\textgreater{}, ``a'', ``d'', ``w'', ``s'': don't move, move left, right, up, or down respectively
\end{itemize}
\noindent
You and the assistant cannot occupy the same location. Your action is replaced by \textless{}Space\textgreater{} in this case.

\end{document}